\documentclass[11pt]{article}

\usepackage[final]{acl}

\usepackage{times}
\usepackage{latexsym}

\usepackage[T1]{fontenc}

\usepackage[utf8]{inputenc}

\usepackage[T1]{fontenc}
\usepackage[most]{tcolorbox}
\usepackage{xcolor}
\usepackage{listings}
\usepackage{bm}
\tcbuselibrary{listings,skins,breakable}
\usepackage[section]{placeins}

\usepackage{enumitem}
\setlist[itemize]{itemsep=2pt, topsep=2pt, parsep=0pt, partopsep=0pt}
\setlist[enumerate]{itemsep=2pt, topsep=2pt, parsep=0pt, partopsep=0pt}
\usepackage{microtype}

\usepackage{inconsolata}

\usepackage{graphicx}
\usepackage[ruled,vlined]{algorithm2e}
\usepackage{multirow}
\usepackage{amsmath} 
\usepackage{booktabs}

\title{KREL: Automatic Medical Coding via Knowledge-Guided Reasoning over Clinical Evidence with LLMs}

\author{
\textbf{Xubin Chen\textsuperscript{1}},
\textbf{Yipeng Zhou\textsuperscript{1}\thanks{Corresponding author.}},
\textbf{Wen Sun\textsuperscript{2}},
\textbf{Chengkai Huang\textsuperscript{3}},\\
\textbf{Xiaoming Fu\textsuperscript{4}},
\textbf{Quan Z. Sheng\textsuperscript{5}}
\\[0.8em]
\textsuperscript{1}Macquarie University,
\textsuperscript{2}Beijing Intelligent Decision Medical Technology Co. Ltd.,\\
\textsuperscript{3}University of New South Wales,
\textsuperscript{4}University of G\"ottingen,\\
\textsuperscript{5}Beijing Normal-Hong Kong Baptist University
\\[0.5em]
\texttt{xubin.chen@students.mq.edu.au},
\texttt{yipeng.zhou@mq.edu.au},
\texttt{sunwen@idmed.cn},\\
\texttt{chengkay.huang@gmail.com},
\texttt{fu@cs.uni-goettingen.de},
\texttt{michaelsheng@bnbu.edu.cn}
}

\begin{document}
\maketitle
\begin{abstract}
Automatic Medical Coding (AMC), which assigns standardized International Classification of Diseases (ICD) codes to clinical notes, is essential for medical reimbursement, quality reporting, and clinical research. Existing pre-trained language model (PLM)-based methods typically formulate AMC as an extreme multi-label classification problem over a predefined code set, while recent large language model (LLM)-based approaches instead frame it as generation or multi-step reasoning. However, key challenges remain, including the extreme length of clinical notes that hinders effective interpretation, the vast ICD label space, and complex coding rules that are not explicitly captured by LLMs. In this work, we propose Knowledge-Guided Reasoning over Clinical Evidence with LLMs (KREL), a framework that leverages LLMs for clinical text understanding and reasoning while integrating selected structured coding relations as external knowledge. This design enables tight coupling between domain knowledge and LLM reasoning, reducing hallucinations and supporting rule-aware verification. Experiments on benchmark datasets show that KREL consistently outperforms strong PLM-based and state-of-the-art LLM-based baselines. Our code is available \href{https://github.com/Progbin/KREL}{here}.

\end{abstract}

\section{Introduction}
Medical coding is the process of assigning relevant diagnosis codes to each clinical note, which is essential in public health and healthcare systems \cite{ji2024unified, gao2024optimising}. This task is generally performed by experienced human annotators who follow the World Health Organization’s ICD coding guidelines when assigning codes. This manual process is notoriously time-consuming and error-prone due to complex rules and the requirement of domain-specific knowledge \cite{yan2022survey, motzfeldt2025code}. 

\begin{figure}
    \centering
    \includegraphics[width=1\linewidth]{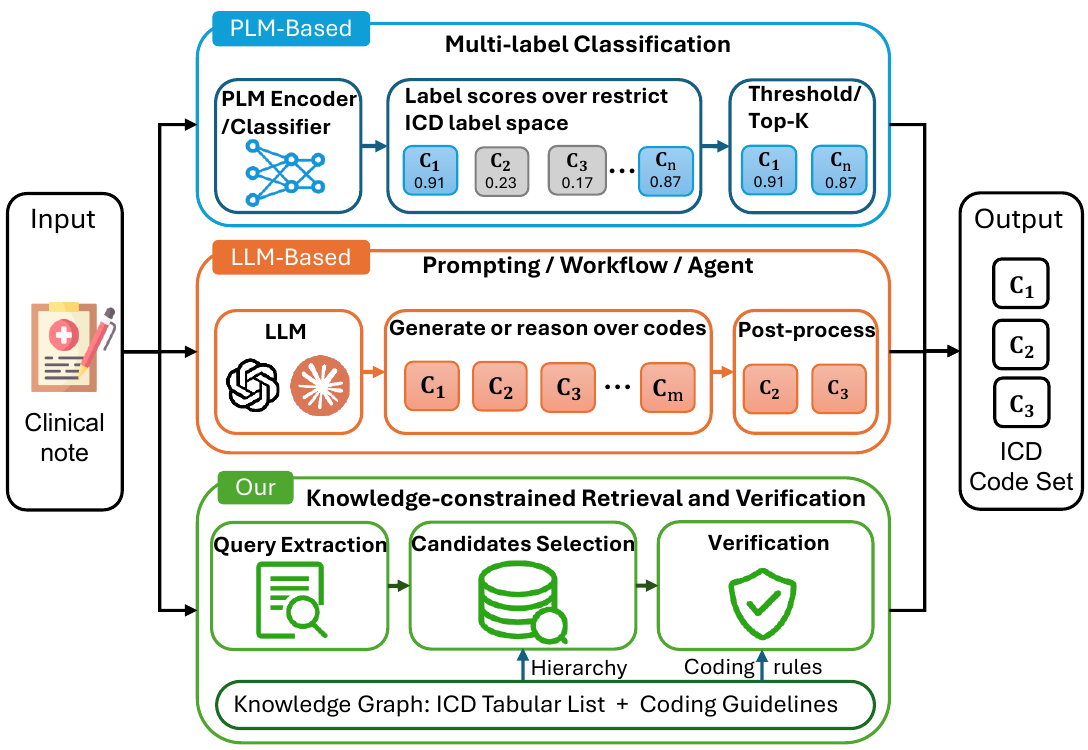}
    \caption{The difference between our framework and existing approaches. Our method better leverages LLM reasoning capabilities by integrating external knowledge.}
    \label{fig:comparework}
\end{figure}

The high cost of medical coding has motivated extensive research on AMC, which aims to automatically assign ICD codes to each clinical note. Existing methods can be broadly categorized into traditional machine learning-based and LLM-based approaches. The former formulates AMC as an extreme multi-label classification problem, where neural encoders or pre-trained language models (PLMs) directly map clinical notes to codes within a predefined label space (typically 50--100 labels, which is much smaller than the full ICD taxonomy of over 70K codes) \cite{huang2022plm, liu2022hierarchical, edin2023automated}. Although these methods achieve strong performance on benchmark datasets, they are typically constrained to a predefined label set, making it difficult to scale to the full and evolving ICD taxonomy.
More recent LLM-based approaches reformulate AMC as a generation or multi-step reasoning task, leveraging prompting, external tools, or coding guidelines to better emulate the workflow of human coders \citep{motzfeldt2025code, zheng2025meddcr, yuan2025toward}.

Despite the superior performance of LLM-based approaches over traditional methods, LLMs alone remain insufficient for AMC due to three key challenges. First, most foundation LLMs are not specifically trained for AMC; even when they can process long clinical notes, they often fail to accurately map clinical concepts to the correct codes. Second, the ICD taxonomy contains over 70K labels, making the label space extremely large and causing LLMs to frequently miss rare or low-frequency (``cold”) codes. Third, ICD coding is governed by complex rules and guidelines, which are not inherently encoded in LLMs, leading to outputs that may violate coding constraints or standards.


These challenges suggest that LLMs should be jointly leveraged with external knowledge to improve AMC accuracy. In light of this, we propose the \textbf{Knowledge-Guided Reasoning over Clinical Evidence with LLMs (KREL)} framework, which consists of three main components. \textbf{Query Extractor:} Given a clinical note, an LLM-based query extractor transforms code-relevant clinical descriptions into structured queries, enabling robust handling of long and unstructured narratives in clinical notes. \textbf{Candidate Selector:} Based on these queries, a guideline-driven knowledge-graph retrieval-augmented generation (KG-RAG) module retrieves and consolidates candidate ICD codes from the full ICD taxonomy, along with their definitions and coding rules. \textbf{Code Verifier:} The clinical note, together with the retrieved candidate codes, definitions, and coding rules, is then fed back into the LLM for verification, where the LLM reviews all evidence to make the final coding decisions. Notably, our design effectively integrates LLM reasoning with external ICD knowledge, enabling scalability to the full ICD-10-CM label space.

Figure~\ref{fig:comparework} compares our framework with existing approaches and highlights its novelty in integrating LLMs with external knowledge for AMC. We present our main contributions as follows:
\begin{itemize}
    \item We propose the novel KREL framework for AMC. Instead of directly generating ICD codes from clinical notes, KREL formulates AMC as a retrieval, candidate refinement, and verification process grounded in structured ICD knowledge and selected structured coding relations.
    
    \item Our framework can scale to the full ICD coding space while leveraging ICD knowledge to guide the code reasoning process. In addition, we demonstrate its interpretability and effectiveness in improving AMC accuracy compared with baseline methods.
    
    
    \item We perform extensive experiments on various datasets, including MDACE, ACI-BENCH, and MIMIC-IV, demonstrating consistent improvements over competitive PLM-based and state-of-the-art LLM-based baselines.
\end{itemize}





\section{Related Work}  

\subsection{Automatic Medical Coding}

Early AMC systems used rule-based and knowledge-driven NLP pipelines, which provided explicit control but required substantial maintenance across institutions and documentation styles \citep{farkas2008automatic,kang2013using}. The release of large EHR datasets such as MIMIC-III and MIMIC-IV helped shift AMC research towards extreme multi-label classification \citep{johnson2016mimic,johnson2023mimic}. Neural models, including CNN- and RNN-based encoders with label-wise attention, improved code-specific evidence modelling \citep{mullenbach2018explainable,vu2020label}. Later work incorporated ICD hierarchy, lexical resources, and domain-adapted PLMs to better handle long notes and fine-grained code distinctions \citep{liu2022hierarchical,mahdi2024co,huang2022plm}. These classification-based methods are effective in restricted or dataset-observed label spaces, but scaling to the complete ICD-10-CM taxonomy remains difficult \citep{huang2022plm,edin2023automated}.

LLM-based methods instead formulate AMC as code generation or multi-step reasoning. Direct prompting with general-purpose LLMs, including ChatGPT and Gemini, remains unreliable under large ICD label spaces because generated codes can be invalid, unsupported, or semantically close but coding-inaccurate \citep{openai2024gpt4technicalreport,geminiteam2025geminifamilyhighlycapable,mustafa2025evaluating}. Recent workflow-based systems, such as Code Like Humans and MedDCR, decompose coding into multiple steps and incorporate ICD resources or coding procedures to improve reliability \citep{motzfeldt2025code,zheng2025meddcr,yuan2025toward}. These approaches move beyond direct prompting, but they still leave open how to combine full-space candidate retrieval, evidence grounding, and rule-aware verification within a single framework. This motivates our retrieval-and-verification formulation for AMC.

\subsection{Knowledge-guided AMC}


Retrieval-augmented generation (RAG) has been used to ground LLM outputs in external evidence for knowledge-intensive and clinical NLP tasks \citep{xiong2024benchmarking,zhao2025medrag,yang2025retrieval}. Standard RAG is less suited to AMC because ICD-10-CM combines a large hierarchical label space with structured tabular conventions and broader narrative coding guidance \citep{cmsnchs2022icd10cm_guidelines,nchs2022icd10cm_tabular}. Candidate construction, therefore, requires both semantic matching and structured coding dependencies.

Knowledge graphs provide a natural way to encode medical concepts and relations, and have been used for entity linking, concept normalization, and clinical reasoning \citep{wu2025medical}. Existing KG-based methods usually focus on concept-level inference rather than full-label-space ICD assignment. Our framework instead uses an ICD knowledge graph to support hierarchy-aware candidate retrieval and rule-aware verification. The hierarchy guides candidate search, while coding-rule relations are converted into verification context for evidence-grounded code selection.

\section{KREL Framework}
\begin{figure*}[t]
    \centering
    \includegraphics[width=\textwidth]{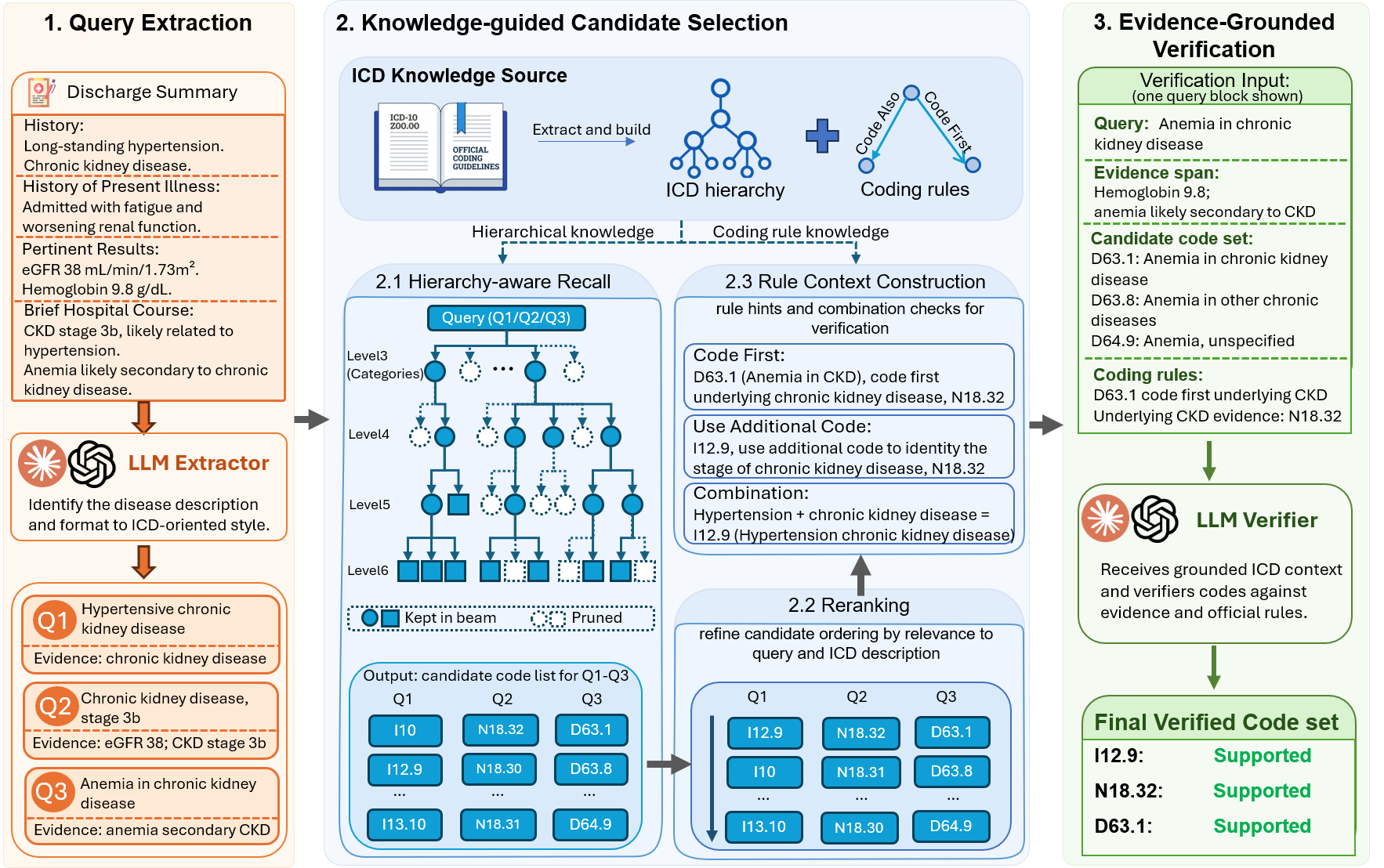}
    \caption{Overview of the KREL framework. (1) The query extractor reads the clinical note, identifies disease descriptions, and reformulates them into ICD-oriented queries (e.g., converting ''CKD stage 3b'' into ''Chronic kidney disease stage 3b''). (2) The selector retrieves candidate ICD codes based on ICD hierarchy knowledge and converts structured coding relations into verification hints. (3) The verifier makes the final coding decisions by jointly considering the candidate codes and all available information and evidence.}
    \label{fig:workflow}
\end{figure*}

\subsection{Problem Formulation}

AMC is commonly formulated as an extreme multi-label classification task, which aims to map a clinical note to a set of ICD codes. Let $x$ denote a clinical note and  $\mathcal{C}$ denote the complete ICD-10-CM code space,\footnote{ICD has multiple versions. This study mainly focuses on ICD-10-CM; adapting the framework to other versions would require rebuilding the corresponding code hierarchy and guideline resources.} which contains over 70K labels. The goal is to predict a code set $\hat{\mathcal{Y}} \subseteq \mathcal{C}$ that matches the gold code set $\mathcal{Y} \subseteq \mathcal{C}$.

In practice, $\mathcal{Y}$ may contain multiple interdependent labels constrained by the hierarchical structure of ICD codes and official coding guidelines. Thereby, AMC is more complex than flat label prediction without considering the interdependent relations between assigned labels. 
In addition, clinical expressions in free-text notes often do not align directly with standardized ICD concepts, which further complicates this problem. 

To address these challenges, we reformulate AMC as a \textbf{hierarchy-aware retrieval and rule-aware verification process}. Instead of directly predicting codes from clinical notes, our framework first extracts evidence-grounded clinical queries and retrieves candidate ICD codes for each query using the ICD hierarchy. It then verifies the retrieved candidate codes against the clinical note, extracted evidence, and coding rules.

\subsection{Framework Overview}


Figure~\ref{fig:workflow} illustrates the KREL framework with a running example. Given a clinical note, the Query Extractor produces evidence-grounded clinical queries. The Candidate Selector retrieves and reranks ICD candidates for each query using ICD hierarchy and code descriptions, and converts coding-rule relations into rule hints and combination checks. The Code Verifier then judges the selected candidates against the full note, localized evidence spans, and rule context to produce the final ICD code set.





\subsection{Query Extractor}

For the query extractor, we first sectionize the clinical note and mark major note sections. The sectioned note is then processed by an LLM, which identifies coding-relevant diagnoses, conditions, symptoms, and clinical status descriptions, and rewrites them into concise queries for subsequent processing. The prompt used for query extraction by the LLM is provided in the Appendix ~\ref{prompt_template}.

Formally, given a clinical note $x$, the query extractor produces a set of query-evidence pairs, represented by
$\mathcal{Q}(x)=\{(q_i,E_i)\}_{i=1}^{n}$, where $q_i$ is the text describing diagnoses, conditions, symptoms, and clinical status, and $E_i$ denotes the associated evidence spans from the clinical note. 



\subsection{Knowledge-guided Candidate Selection}

The selector maps each extracted query to ICD candidate lists and builds rule context for verification. It uses an ICD knowledge graph constructed from ICD code text, hierarchy, and guideline-derived relations.

\subsubsection{ICD Knowledge Graph Construction}

We construct an ICD knowledge graph offline to encode the ICD knowledge used by downstream stages. The graph contains three types of information. First, each ICD code is associated with textual attributes, including its official description and inclusion terms, which support query--code matching. Second, ICD-10-CM provides a multi-level hierarchy, where broad categories are refined into more specific codes. Third, ICD coding guidelines define rule relations, including pairwise coding-rule relations such as \textit{codeFirst}, \textit{useAdditionalCode}, and \textit{codeAlso}, and multi-code combination-code relations where required codes jointly support a target combination code.

Formally, we represent the graph as \(\mathcal{G}_{\mathrm{ICD}} = (\mathcal{V}_{C}, \mathcal{E}_{H}, \mathcal{E}_{R}, \mathcal{E}_{M}),\) where $\mathcal{V}_{C}$ denotes ICD code vertices, $\mathcal{E}_{H}$ denotes hierarchical relations, $\mathcal{E}_{R}$ denotes pairwise coding-rule relations, and $\mathcal{E}_{M}$ denotes multi-code combination-code relations. Each code $c_v\in\mathcal{V}_{C}$ is associated with textual attributes $d_v$, which are used for retrieval and reranking.

These graph components are used at different stages. The hierarchy $\mathcal{E}_{H}$ defines the search space for hierarchy-aware recall, while the code text $d_v$ supports semantic scoring and reranking. The rule relations $\mathcal{E}_{R}$ and $\mathcal{E}_{M}$ are not used to expand the ordinary candidate lists. They are converted into rule hints and combination checks before verification. The full KG schema is provided in Appendix~\ref{app:kg-schema}.

\subsubsection{Hierarchy-aware Candidate Recall}


Given a query $q_i$, the recall stage searches the ICD code hierarchy along $\mathcal{E}_{H}$ 
to obtain a core candidate set, denoted by $\mathcal{R}_i$. 
The relevance between $q_i$ and code $c$ is computed from their embeddings, where we adopt the same embedding LLM to generate embeddings for $q_i$. The instruction used for the embedding LLM is provided in Appendix ~\ref{prompt_template}.
Formally, let $\mathbf{z}_{q_i}$ and $\mathbf{z}_{c}$ denote the embeddings of $q_i$ and code $c$. 
Then, the relevance score between $q_i$ and $c$ is $r_{emb}(q_i,c)=\mathbf{z}_{q_i}^{\top}\mathbf{z}_{c}.$


Based on the relevance score, we retrieve ICD codes related to a query using the Hierarchy-Aware Beam Search (HBS) algorithm. To control the search cost, the retrieval process is constrained by a set of budget hyper-parameters $K_c$, $B$, $M$, $D$, and $K_f$, where $K_c$ denotes the number of initial search codes, $B$ is the beam size, $M$ is the maximum number of child codes explored for each parent code, $D$ is the maximum search depth, and $K_f$ is the maximum number of final selected codes.

Specifically, HBS starts from the top-$K_c$ ICD nodes that are most relevant to  $q_i$, ranked in descending order according to $r_{emb}(q_i, c)$. The algorithm then progressively descends along the ICD hierarchy to search for more specific child codes. For each selected parent node with child nodes, HBS retains the top-$M$ child codes with the highest scores, where the scoring function is defined later. At each level (i.e., search depth), all candidate child nodes are globally sorted, and only the top-$B$ nodes are preserved as the beam for the next search step.

For a child code $c$ expanded from its parent code $c'$, its score is computed as a combination of the parent score and the local relevance of the child node, i.e.,
$
S(c) = \lambda S(c') + (1 - \lambda) r_{emb}(q_i, c),
$
to capture information propagated along the search path. In our implementation, we set $\lambda = 0.5$. Note that a leaf node may be reached via multiple paths; in such cases, we retain only the highest score among all paths. Finally, we select the top-$K_f$ leaf nodes with the highest scores to form $\mathcal{R}_i$. Details of the HBS algorithm are provided in Appendix~\ref{app:hbs}.

It is worth noting that HBS can outperform existing greedy-based retrieval algorithms \cite{boyle2023automated} in terms of recall, as it defers code pruning to leaf nodes. In contrast, greedy-based methods prune parent codes at earlier stages, which may lead to the premature removal of correct codes.



\subsubsection{Reranking}

The candidate set produced by HBS, i.e., $\mathcal{R}_i$, still contains noisy codes, such as sibling codes or broadly related codes with similar descriptions. To filter out these noisy candidates, we rerank and subsequently truncate $\mathcal{R}_i$.

For reranking, we employ a cross-encoder LLM with pairwise relevance scoring. Specifically, for each $c \in \mathcal{R}_i$, we extract its ICD code description, including symptoms, concepts, and inclusion terms, from the KG. A pairwise reranker LLM is then used to assign a relevance score to each query--code pair $(q_i, c)$ based on the extracted code description. The instruction used for the LLM reranker to generate relevance score is presented in Appendix \ref{prompt_template}.  All candidate codes are subsequently sorted in descending order according to their relevance scores  $r_{rerank}(q_i, c)$. Let $\tilde{\mathcal{L}}_i$ denote the reranked list of candidate codes.



Since $K_f$ is typically large, we further apply a note-level verification budget $K_V$ to reduce the cost of subsequent verification. For a note with $n$ extracted queries, we distribute this budget approximately evenly across the reranked lists $\{\tilde{\mathcal{L}}_i\}_{i=1}^n$ and retain the top-ranked codes from each list. If some queries contain fewer candidates than their allocated budget, the unused slots are reassigned to the highest-ranked remaining candidates across all queries. We denote the resulting budgeted candidate list for query $q_i$ as $\mathcal{L}_i$.



\subsubsection{Augmenting Candidate Codes with Rules}


Until now, candidate codes have been selected solely by utilizing $\mathcal{E}_{H}$ in the KG. We further incorporate the knowledge encoded in $\mathcal{E}_{R}$ and $\mathcal{E}_{M}$ to support rule-aware verification, focusing only on coding relations that can be represented as explicit pairwise or multi-code combination dependencies.


According to the ICD coding guidelines, we identified three types of rule relations: \textit{codeFirst}, \textit{useAdditionalCode}, and \textit{codeAlso}. We focus on these relations because they can be automatically extracted and represented as explicit structured dependencies between ICD codes. Among them, the first two relation types are considered more critical.
For a code $c \in \mathcal{L}_i$, we retrieve the top $K_a$ related codes $c'$ such that $c$ and $c'$ are connected through either \textit{codeFirst} or \textit{useAdditionalCode} relations. Here, both $c$ and $c'$ are selected for the same clinical note, although they may belong to different candidate lists $\mathcal{L}_i$. The retrieved codes are prioritized according to their reranking scores obtained in the reranking step.
If fewer than $K_a$ related codes are retrieved, we further retrieve additional codes connected through the \textit{codeAlso} relations by repeating the above process.

For each selected $c'$, we generate a textual description of the relation between $c$ and $c'$ as a relation hint. We aggregate all such relation hints to form a hint set, denoted as $H_c$. Let $\mathcal{H}_i$ denote the rule hint set of all $H_c$ from query $q_i$.  Subsequently, $\mathcal{H}_i$ is used together with $q_i$ to facilitate the final code assignment. For example, Figure~\ref{fig:workflow} shows how the structured relation for D63.1 is converted into the verification hint ``D63.1 code first underlying CKD,'' together with N18.32 as the supported underlying condition.


Next, we consider the knowledge encoded in $\mathcal{E}_{M}$. Each relation in $\mathcal{E}_{M}$ specifies the relationship between a set of required codes and a target combination code. Since the required codes may be distributed across different queries, we construct the aggregated candidate set for clinical note $x$ as $\mathcal{L}_x = \bigcup_{i=1}^{n_x} \mathcal{L}_i$. If all required codes associated with a combination code in $\mathcal{E}_{M}$ are present in $\mathcal{L}_x$, we include the corresponding combination rule, linking the required codes to the combination code, in the set $\Omega_x$.


\subsection{Evidence-grounded Code Verification}

We employ an LLM verifier to perform the final code assignment based on all the collected information. For each clinical note $x$, we invoke the  LLM  verifier once to predict codes using the inputs $x$, $q_i$, $E_i$, $\mathcal{L}_i$, and $\mathcal{H}_i$ for $1 \leq i \leq n$, as well as $\Omega_x$, following a two-step process. The instruction used for the LLM verifier is presented in Appendix \ref{prompt_template}.

First, for each query $q_i$, the  LLM verifier compares candidate codes in $\mathcal{L}_i$ against the full clinical note, the localized evidence spans $E_i$, and the corresponding code descriptions. Multiple codes may be selected for a single query, and it is also possible that no code is supported by a given query. When a candidate code is associated with any rule hint in $\mathcal{H}_i$, the verifier further evaluates the corresponding coding dependencies across all queries within $x$. For each verified code, the verifier LLM assigns one of the following ratings: \textsc{supported}, \textsc{possible}, or \textsc{not supported}. 

Second, after processing queries, we verify combination rules in $\Omega_x$. For each remaining combination rule, we send both the combination code, its required component codes, the clinical note and other information to the verifier, who will decide whether to support this combination code for $x$.

Finally, all codes rated as \textsc{supported} or \textsc{possible} are retained and included in the set $\hat{\mathcal{Y}}$ as prediction codes for the clinical note $x$.

\section{Experimental Setup}
\subsection{Datasets}

We evaluate our framework on three clinical datasets: MDACE \citep{cheng2023mdace}, ACI-Bench \citep{yim2023aci}, and MIMIC-IV \citep{johnson2023mimic}. MDACE is a manually annotated medical coding dataset with ICD code assignments and code-related annotation information. ACI-Bench provides clinical notes with sentence-level links between text spans and ICD codes, supporting both performance evaluation and evidence-based analysis. MIMIC-IV is a large-scale de-identified hospital EHR database containing discharge summaries and ICD-coded diagnoses.

For MDACE and ACI-Bench, we use the official dataset splits and label inventories provided by the benchmark authors to ensure direct comparability with reported baselines.

We evaluate two label-space settings. In the \textit{benchmark-label-space setting}, predictions are constrained to the predefined ICD label inventory of each benchmark. This setting is used for MDACE and ACI-Bench. In the \textit{full-label-space setting}, methods select codes from the complete ICD-10-CM code space with over 70K labels. This setting is used for the fully labelled version of MDACE and a randomly sampled subset of 500 MIMIC-IV. The MIMIC-IV-500 is used as a computationally tractable feasibility evaluation under the complete ICD-10-CM label space. We further compare its aggregate code-frequency statistics with those of the full eligible MIMIC-IV corpus and observe similar distributions; detailed sampling procedures and statistics are provided in Appendix~\ref{app:impl:dataset}.

\subsection{Baselines}
We compare our framework with two categories of baselines: \textbf{PLM-based methods} and \textbf{LLM-based methods}. 

PLM-based methods formulate AMC as a supervised extreme multi-label classification task. In this category, \textbf{PLM-ICD} \cite{huang2022plm} and \textbf{PLM-CA} \cite{douglas2025less} are representative baselines in the previous AMC study. They perform code prediction over a predefined label space and use label-wise designs to better capture note--code correspondence, providing limited interpretability and auditability.

Instead, LLM-based methods formulate AMC as a generation or reasoning task. Existing works, such as \textbf{GPT-4o} \cite{openai2024gpt4ocard}, \textbf{CoT} \cite{wei2022chain}, and \textbf{CoT-SC} \cite{wang2022self}, rely on direct prompting or generic reasoning strategies to generate codes from clinical notes directly. Others, including \textbf{RRS} \cite{kwan2024large}, \textbf{MAC} \cite{li2024exploring}, \textbf{CLH} \cite{motzfeldt2025code}, and \textbf{MedDCR} \cite{zheng2025meddcr}, model AMC as a multi-step workflow and leverage LLMs together with external tools or coding resources to better simulate the human coding process for AMC. 


\begin{table*}[t]
\centering
\small
\begin{tabular}{llcccccc}
\toprule
\multicolumn{8}{l}{\textbf{(a) Benchmark-label-space setting}} \\
\midrule
\multirow{2}{*}{Method category} &
\multirow{2}{*}{Model} &
\multicolumn{3}{c}{MDACE (CM + PCS)} &
\multicolumn{3}{c}{ACI-BENCH} \\
\cmidrule(lr){3-5} \cmidrule(lr){6-8}
 & & Precision & Recall & F1 & Precision & Recall & F1 \\
\midrule
\multirow{2}{*}{PLM}
 & ICD    & \textbf{0.49} & 0.47 & 0.48 & 0.43 & 0.41 & 0.42 \\
 & CA     & 0.46 & 0.45 & 0.45 & 0.44 & 0.42 & 0.43 \\
\midrule
\multirow{2}{*}{LLM prompting}
 & CoT    & 0.30 & 0.31 & 0.30 & 0.35 & 0.50 & 0.41 \\
 & CoT-SC & 0.39 & 0.43 & 0.41 & 0.36 & 0.59 & 0.44 \\
\midrule
\multirow{4}{*}{LLM workflow}
 & RRS    & 0.24 & 0.30 & 0.27 & 0.26 & 0.52 & 0.35 \\
 & MAC    & 0.27 & 0.31 & 0.29 & 0.23 & 0.50 & 0.31 \\
 & CLH    & 0.45 & 0.40 & 0.42 & 0.44 & 0.39 & 0.41 \\
 & MedDCR & 0.41 & 0.55 & 0.47 & 0.43 & 0.67 & 0.52 \\
\midrule

& KREL   & 0.42 & \textbf{0.59} & \textbf{0.49} 
          & \textbf{0.61} & \textbf{0.82} & \textbf{0.70} \\

\midrule
\addlinespace[2pt]
\multicolumn{8}{l}{\textbf{(b) Full-label-space setting}} \\
\midrule
\multirow{2}{*}{Method category} &
\multirow{2}{*}{Model} &
\multicolumn{3}{c}{MDACE (CM only)} &
\multicolumn{3}{c}{MIMIC-IV-500} \\
\cmidrule(lr){3-5} \cmidrule(lr){6-8}
 & & Precision & Recall & F1 & Precision & Recall & F1 \\
\midrule
\multirow{3}{*}{LLM prompting}
 & GPT-4o & 0.38 & 0.25 & 0.30 & 0.47 & 0.24 & 0.31 \\
 & CoT      & 0.40 & 0.26 & 0.31 & \textbf{0.49} & 0.23 & 0.32 \\
 & CoT-SC   & 0.39 & 0.26 & 0.32 & \textbf{0.49} & 0.25 & 0.33 \\
\midrule
LLM workflow
 & CLH      & 0.40 & 0.23 & 0.29 & 0.44 & 0.18 & 0.25 \\
\midrule

& KREL     & \textbf{0.48} & \textbf{0.53} & \textbf{0.51} 
            & 0.45 & \textbf{0.35} & \textbf{0.39} \\
\bottomrule
\multicolumn{8}{l}{\footnotesize \textit{Note}: Baseline results in Panel (a) are reported by \citet{zheng2025meddcr}.}
\end{tabular}
\caption{Performance comparison under benchmark-label-space and full-label-space settings}
\label{tab:main_results_combined}
\end{table*}

\subsection{Implementation Details}
We instantiate our framework with \textbf{GPT-4o} \cite{openai2024gpt4ocard} as the backbone LLMs, which are used for both the query extractor and verifier. We construct the knowledge graph from the \textbf{ICD-10-CM Tabular List} \cite{nchs2022icd10cm_tabular} and the corresponding \textbf{Official Coding Guidelines} (2022 version) \cite{cmsnchs2022icd10cm_guidelines}. We employ \textbf{Qwen3-Embedding-8B} \cite{zhang2025qwen3embeddingadvancingtext}  as the embedding LLM and \textbf{Qwen3-Reranker-8B} \cite{zhang2025qwen3embeddingadvancingtext} as the reranking LLM in Candidate Selector. In the HBS algorithm, we set $K_c$=200, $B$=200, $M$=8, $D$=6 and $K_f$=30. After deduplicating candidates across queries, we keep at most $K_V = 50$ codes per note for final verification.



\subsection{Evaluation Metrics}
Following previous work \cite{zheng2025meddcr, motzfeldt2025code, huang2022plm} in automatic medical coding, we evaluate all methods using precision, recall and F1-score, and primarily report \textbf{micro-averaged} results. 
Although our method reformulates AMC as a retrieval-and-verification problem rather than a standard multi-label classification task, the final outputs are converted into the same encounter-level ICD code predictions used in previous work, ensuring direct and fair comparison.

\section{Results and Analysis}

\subsection{Main Results}

Table~\ref{tab:main_results_combined} reports results under both benchmark-label-space and full-label-space settings. Panel (a) shows that KREL remains competitive in the benchmark setting. It achieves the best F1 on both datasets, reaching 0.49 on MDACE and 0.70 on ACI-BENCH. The MDACE gain is modest and mainly recall-driven, while the result on ACI-BENCH shows higher gains in both precision and recall over PLM, prompting, and workflow baselines.

Panel (b) reports the results under the full-label-space setting, which better reflects the main challenge addressed in this work. On MDACE, KREL improves F1 from the best baseline score of 0.32 to 0.51, with recall increasing from 0.26 to 0.53. On MIMIC-IV-500, KREL also obtains the best F1, improving from 0.33 to 0.39. KREL improves over the strongest baseline on both full-label-space datasets, with a larger gain observed on MDACE. These results indicate that retrieval and verification over structured ICD knowledge is more effective than direct prompting or workflow baselines when the full-label space is considered.

\subsection{Ablation Study}
\begin{table}[t]
\centering
\small
\begin{tabular}{llccc}
\toprule
Variant & Type & Precision & Recall & F1 \\
\midrule
KREL (full) & -- & \textbf{0.48} & 0.53 & \textbf{0.51} \\
medSpaCy query & Replace & 0.35 & 0.19 & 0.24 \\
flat retrieval & Replace & 0.46 & 0.50 & 0.48 \\
w/o reranker & Remove & 0.46 & 0.46 & 0.46 \\
w/o rule injection & Remove & 0.47 & 0.50 & 0.49 \\
w/o verifier & Remove & 0.11 & \textbf{0.62} & 0.19 \\
\bottomrule
\end{tabular}
\caption{Ablation study on MDACE under the full-label-space setting.}
\label{tab:ablation}
\end{table}

Table~\ref{tab:ablation} reports ablation results on MDACE under the full-label-space setting, where we separately remove or replace each critical component in our framework.

Since removing the query extractor entirely causes system collapse, we replace the LLM-based query extractor with the NER module from medSpaCy, which extracts medical terms from clinical notes and uses them directly as retrieval queries without evidence-grounded query reformulation. This replacement significantly degrades performance, reducing recall from 0.53 to 0.19 and F1 from 0.51 to 0.24, indicating that evidence-grounded query construction is crucial for achieving adequate candidate coverage.

\begin{table}[t]
\centering
\small
\setlength{\tabcolsep}{4pt}
\begin{tabular}{lrrrr}
\toprule
\textbf{Rule} & \textbf{GT (share)} & \textbf{KREL} & \textbf{CoT-SC} & \textbf{CLH} \\
\midrule
Pairwise    & 17 (3.14\%) & 12 / 70.6\% & 5 / 29.4\% & 2 / 11.8\% \\
Combination & 13 (2.40\%) & 8 / 61.5\%  & 1 / 7.7\%  & 1 / 7.7\% \\
\midrule
All         & 30 (5.55\%) & 20 / 66.7\% & 6 / 20.0\% & 3 / 10.0\% \\
\bottomrule
\end{tabular}
\caption{Performance on rule-dependent ground-truth note--code pairs in MDACE. Model columns report correct predictions / recall.}
\label{tab:rule_dependent}
\end{table}

We further examine the effect of removing the verifier, which leads to a substantial drop in precision from 0.48 to 0.11, despite an increase in recall to 0.62. This demonstrates that the verifier is essential for filtering unsupported retrieved candidates. Removing the reranker reduces  F1 to 0.46, while removing coding rules slightly decreases F1 to 0.49. The modest aggregate F1 change reflects the low prevalence of explicitly rule-dependent targets. Under our broad definition, only 30 of 541 ground-truth note–code assignments (5.55\%) are rule-dependent. Within this subset, KREL recovers 20/30 targets, compared with 6/30 for CoT-SC and 3/30 for CLH; detailed results by rule type are reported in Table ~\ref{tab:rule_dependent}. Overall, these results show that each component contributes meaningfully to the overall performance of the framework. 

\subsection{Retrieval Strategy Analysis}


\begin{figure}
    \centering
    \includegraphics[width=1\linewidth]{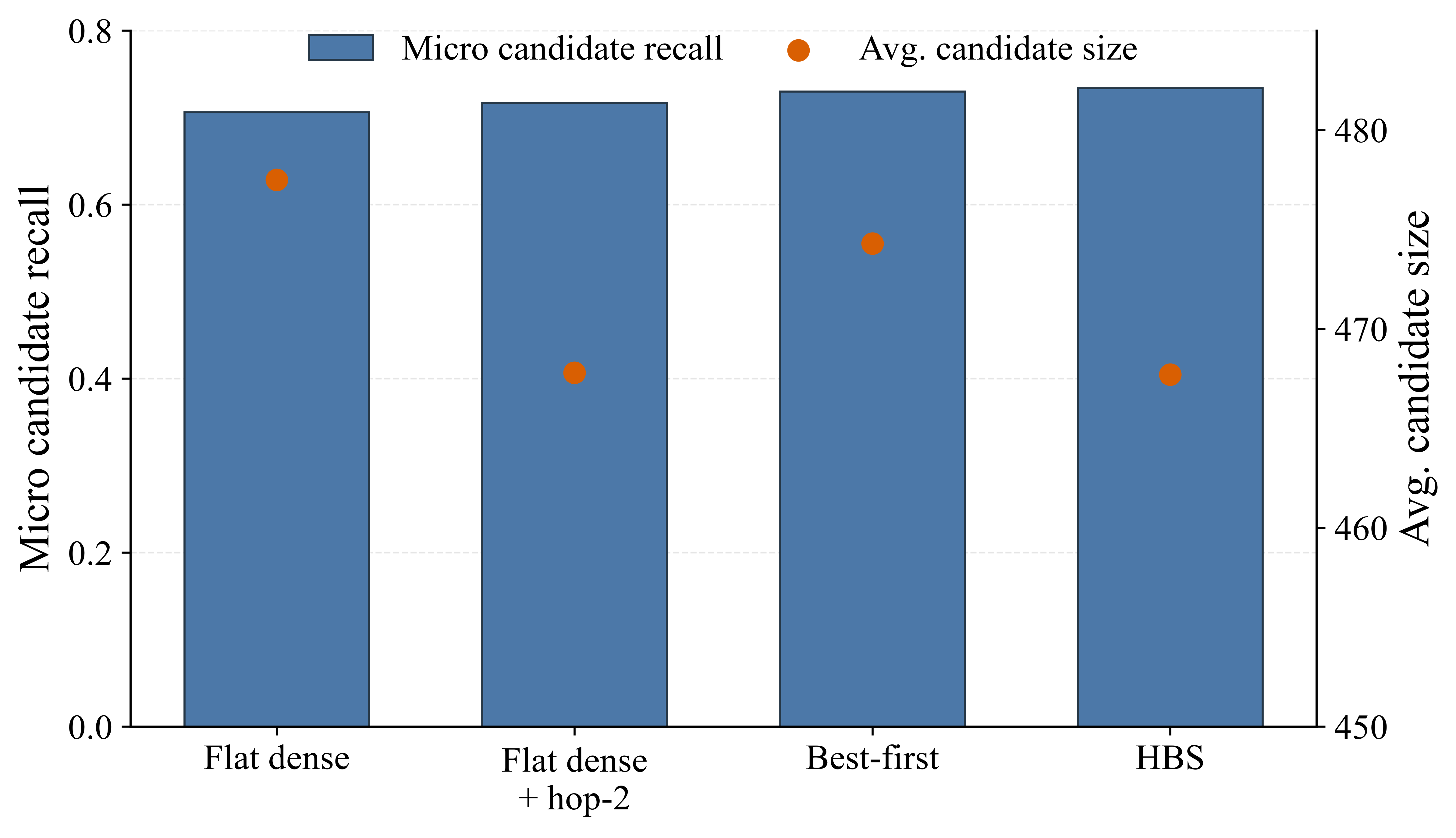}
    \caption{Comparison of different strategies for candidate code retrieval on MDACE under the full-label-space setting.}
    \label{fig:retrieval_strategy}
\end{figure}

Figure~\ref{fig:retrieval_strategy} compares HBS with alternative retrieval strategies for candidate generation on MDACE under the full-label-space setting. In addition to exact flat dense retrieval, which ranks ICD codes directly by query–code similarity without hierarchical traversal, we consider two stronger alternatives: flat dense retrieval followed by two-hop graph expansion, and best-first hierarchical search using the same relevance scorer. Flat dense retrieval achieves a micro candidate recall of 0.706 with 477.5 candidates per note. Two-hop graph expansion improves recall to 0.717 while using 467.8 candidates, indicating that incorporating local graph structure provides additional coverage. Best-first hierarchical search further increases recall to 0.730 with 474.3 candidates. HBS achieves the highest recall of 0.734 while using 466.7 candidates per note, yielding the best observed recall–candidate-budget trade-off among the evaluated strategies. These results suggest that the gains of HBS arise not merely from accessing the ICD graph, but from structured hierarchical traversal that preserves multiple promising branches while controlling candidate growth.

\subsection{Candidate Set Size Evolution}

\begin{figure}
    \centering
    \includegraphics[width=1\linewidth]{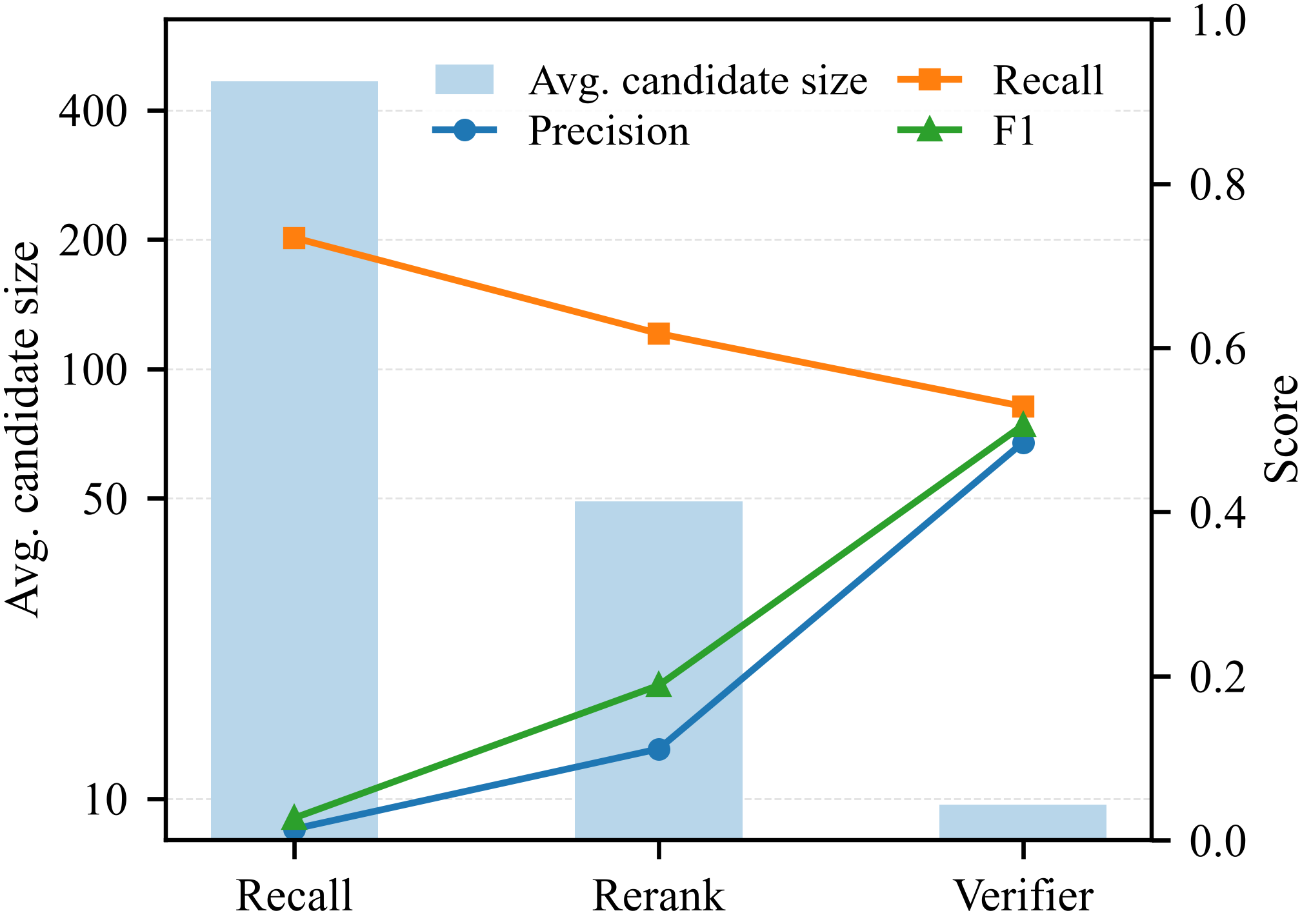}
    \caption{Evolution of the candidate set size against recall and precision performance.}
    \label{fig:pipeline_evolution}
\end{figure}

We further empirically analyze how the size of the candidate code set evolves across the framework pipeline.
Figure~\ref{fig:pipeline_evolution} illustrates how the size evolves on MDACE under the full-label-space setting. The recall stage by retrieving candidate codes via HBS favours coverage with 467 codes per note on average. The average recall is 0.734, but its precision remains low at 0.01. Reranking reduces the candidate size to 49 and improves precision to 0.11 and F1 to 0.19, while retaining a recall of 0.62. The final verifier further reduces the output size to 10 codes on average and increases precision to 0.48 and F1 to 0.51, with recall decreasing to 0.53. The result indicates that candidate retrieval mainly provides coverage, while reranking and verification improve the reliability of the final code assignment.

\section{Conclusion}

In this work, we proposed \textbf{KREL}, a Knowledge-Guided Reasoning framework over Clinical Evidence with LLMs for automated medical coding. Unlike existing PLM-based methods that treat AMC as an extreme multi-label classification task and recent LLM-based approaches that rely on generation or multi-step reasoning, KREL explicitly integrates selected structured coding relations into the reasoning process of LLMs. This design enables more effective interpretation of long and complex clinical narratives, better handling of the large ICD label space, and improved adherence to coding rules that are not inherently captured by LLMs. By tightly coupling external medical knowledge with LLM reasoning, KREL reduces hallucinations and enhances coding reliability. Extensive experiments on benchmark datasets demonstrate that KREL consistently outperforms strong PLM-based and state-of-the-art LLM-based baselines, showing its effectiveness and robustness for AMC.

\section*{Limitations}

\noindent\textbf{Candidate recall.}
KREL verifies codes from retrieved candidates, so missed candidates cannot be recovered by later stages except for explicitly triggered combination-code checks. Maintaining high candidate recall under a fixed verifier budget remains challenging in the full ICD-10-CM label space, especially for rare or highly specific codes.


\noindent\textbf{Reliance on LLMs.}
Our framework relies on LLMs for query extraction and verification \cite{ye2026memweaver,huang2025towards,cong2026seeing}. While this enables flexible evidence-grounded reasoning, it introduces cost, latency, and reproducibility considerations. Clinical deployment may require secure inference environments, locally hosted LLMs, or institution-approved LLM services.

\noindent\textbf{Data Scarcity.}
Our evaluation is conducted on public EHR datasets using offline metrics. MDACE, ACI-Bench, and MIMIC-IV provide complementary benchmarks, but they do not capture the full diversity of institutional documentation styles or coding workflows. In addition, evidence annotations are not available for all datasets, limiting systematic evaluation of grounding quality. Future work should extend evaluation to more diverse clinical datasets and incorporate feedback from professional medical coders.


\section*{Ethical Considerations}

This work uses publicly available or credentialed-access clinical datasets for automated medical coding research. MDACE and ACI-Bench are public benchmark datasets, while MIMIC-IV is accessed under PhysioNet credentialing and used within the authorised data-use scope. Since AMC is a high-stakes clinical documentation and billing task, the proposed system should not be used as a replacement for professional medical coders or clinicians. Its outputs are intended to support coding review and should be subject to human validation before any clinical, administrative, or billing use. Deployment in real clinical environments would also require secure inference infrastructure, institution-approved models or services, and safeguards to prevent protected health information from being exposed to unauthorised systems.

\section*{Acknowledgments}
This work was partially supported by DAAD23 (Nos. 57692192 and 57702286).


\bibliography{custom}

\appendix
\section*{Appendix}

\section{Experimental Environment}

All local computation, including embedding-based retrieval, candidate
generation, reranking, post-processing, and metric computation, was
conducted on a Linux server running \texttt{Ubuntu 22.04.5 LTS}
with kernel \texttt{Linux 5.15.0-78-generic} and \texttt{glibc 2.35}.
The software stack was based on \texttt{Python 3.12.3} and
\texttt{PyTorch 2.8.0} compiled with \texttt{CUDA 12.8}
(\texttt{torch 2.8.0+cu128}), with \texttt{cuDNN 9.10.2} enabled.

The hardware configuration consisted of one
\texttt{NVIDIA GeForce RTX 5090} GPU with approximately
\texttt{32GB} memory.

Embedding construction, dense retrieval, and reranking were executed with
GPU acceleration on this server. LLM-based query extraction and verifier
stages were executed through API-based batch inference using the
corresponding model providers.

\section{Implementation Details}\label{app:impl}

\begin{table*}[h]
\centering
\small
\begin{tabular}{lllcc}
\toprule
Dataset & Label-space setting (size) & Code type & Test clinical notes & Test unique GT codes \\
\midrule
MDACE & Benchmark (350) & ICD-10-CM + PCS & 122 & CM: 312; PCS: 38 \\
ACI-Bench & Benchmark (132) & ICD-10-CM & 120 & 132 \\
MDACE & Full (72750) & ICD-10-CM & 61 & 282 \\
MIMIC-IV-500 & Full (72750) & ICD-10-CM & 500 & 1,812 \\
\bottomrule
\end{tabular}
\caption{Dataset and evaluation scope.}
\label{tab:dataset_scope}
\end{table*}

\subsection{Dataset and Evaluation Scope}
\label{app:impl:dataset}

Table~\ref{tab:dataset_scope} summarizes the datasets and evaluation scope used in our experiments. MDACE and ACI-Bench are publicly available benchmark datasets derived from MIMIC clinical records, with additional task-specific annotation for medical coding evaluation \citep{cheng2023mdace,yim2023aci}. MIMIC-IV is a credentialed-access clinical database distributed through PhysioNet \citep{johnson2023mimic}. We accessed and used MIMIC-IV after obtaining PhysioNet authorization, and all experiments were conducted within the permitted data-use scope.

\begin{table}[t]
\centering
\caption{Statistics of MIMIC-IV-500 and the full MIMIC-IV pool.
Rare GT denotes assignments whose code frequency in the full pool is $\leq 10$.}
\label{tab:mimiciv_dataset_statistics}
\small
\setlength{\tabcolsep}{5pt}
\begin{tabular}{lrr}
\toprule
\textbf{Statistic} & \textbf{MIMIC-IV} & \textbf{MIMIC-IV-500} \\
\midrule
Clinical notes
& 122,288 & 500 \\

Mean GT codes/note
& 14.435 & 14.894 \\

Median GT codes/note
& 13 & 14 \\

GT assignments
& 1,765,181 & 7,447 \\

Unique GT codes
& 16,155 & 1,812 \\

Rare GT share
& 1.77\% & 1.84\% \\

Top-20 GT coverage
& 22.24\% & 21.66\% \\
\bottomrule
\end{tabular}
\end{table}

For the benchmark-label-space setting, evaluation is restricted to the predefined label inventory of each benchmark. In particular, MDACE uses both ICD-10-CM and ICD-10-PCS labels to align with the baseline setting in \citet{zheng2025meddcr}. For the full-label-space setting, we evaluate over the complete ICD-10-CM diagnosis space, which contains 72,750 terminal diagnosis codes, using MDACE and MIMIC-IV-500. MIMIC-IV-500 consists of 500 discharge summaries randomly sampled from the full MIMIC-IV pool. As shown in Table~\ref{tab:mimiciv_dataset_statistics}, it closely matches the full pool in code density and aggregate code-frequency characteristics, supporting its use as a computationally tractable feasibility benchmark for full-label-space evaluation.

\subsection{Graph Construction}\label{app:kg-schema}
We construct the ICD knowledge graph from ICD-10-CM code metadata, hierarchy information, and coding-rule annotations. The graph is defined as
\[
\mathcal{G}_{\mathrm{ICD}}
=
(\mathcal{V}_{C}, \mathcal{E}_{H}, \mathcal{E}_{R}, \mathcal{E}_{M}),
\]
where $\mathcal{V}_{C}$ is the set of ICD code nodes, $\mathcal{E}_{H}$ is the set of hierarchy edges, $\mathcal{E}_{R}$ is the set of pairwise coding-rule edges, and $\mathcal{E}_{M}$ is the set of multi-code combination relations.

\noindent\textbf{Code nodes}
For each ICD-10-CM code $c$, we create one code node $v_c \in \mathcal{V}_{C}$. Each code node stores the code string, official description, inclusion terms, and version information. We denote the textual attributes of code $c$ as $d_c$.

\noindent\textbf{Hierarchy edges}
For each parent-child relation in the ICD taxonomy, we add a directed edge from the child code to its parent code. These edges form $\mathcal{E}_{H}$. For example, if code $c$ is a more specific code under parent code $p$, the graph contains an edge $c \rightarrow p$.

\noindent\textbf{Pairwise coding-rule edges}
For pairwise coding-rule annotations, we add directed edges between ICD code nodes. Each edge is represented as
\[
(c,\tau,c') \in \mathcal{E}_{R},
\]
where $c$ is the source code, $c'$ is the referenced code, and $\tau$ is the rule type. We include three rule types:
\[
\tau \in \{\textit{codeFirst}, \textit{useAdditionalCode}, \textit{codeAlso}\}.
\]
When a rule annotation refers to a code range, we expand the range into the corresponding ICD codes and create one edge for each target code.

\noindent\textbf{Multi-code combination relations}
Some coding rules require multiple codes before a target combination code can be considered. We store each such rule as a combination relation
\[
m=(\mathrm{Req}(m),\mathrm{Tar}(m)) \in \mathcal{E}_{M},
\]
where $\mathrm{Req}(m)$ is the set of required ICD codes and $\mathrm{Tar}(m)$ is the target combination code. In the graph implementation, each combination relation is stored as a rule node connected to its required code nodes and target code node. 

\subsection{HBS Algorithm} \label{app:hbs}

\section{Experiment}\label{app:experiment}



\subsection{Detailed Evaluation Metric}
\label{app:experiment:eval}

We report micro-averaged precision, recall, and F1 over all notes.
Precision and recall are then defined as
\[
\mathrm{Precision}=\frac{\mathrm{TP}}{\mathrm{TP}+\mathrm{FP}},
\]
and
\[
\mathrm{Recall}=\frac{\mathrm{TP}}{\mathrm{TP}+\mathrm{FN}}.
\]
The F1 score is the harmonic mean of precision and recall:
\[
\mathrm{F1}
=
2\cdot
\frac{\mathrm{Precision}\cdot\mathrm{Recall}}
{\mathrm{Precision}+\mathrm{Recall}}.
\]

In AMC, precision measures how many predicted ICD codes are correct, while recall measures how many gold ICD codes are recovered. F1 provides a single measure of the trade-off between assigning accurate codes and covering all relevant diagnoses. This is important because clinical notes may contain multiple reportable conditions, and missing a relevant code or assigning an unsupported code can both reduce coding quality.

\begin{table*}[t]
\centering
\small
\begin{tabular}{lccc}
\toprule
Metric & LLM prompting & LLM-agent (CLH) & KREL \\
\midrule
Average predicted codes per note & 5.98 & 5.11 & 9.67 \\
False positives & 222 & \textbf{187} & 304 \\
False negatives & 398 & 416 & \textbf{255} \\
Combination-code recall & 0.077 & 0.077 & \textbf{0.615} \\
Sibling-code FP rate & 0.329 & \textbf{0.219} & 0.227 \\
\bottomrule
\end{tabular}
\caption{Comparison of error patterns across method families on MDACE ICD-10-CM discharge summaries under the full-label-space setting.}
\label{tab:method_family_error_patterns_full_label}
\end{table*}

\begin{figure*}
    \centering
    \includegraphics[width=\textwidth]{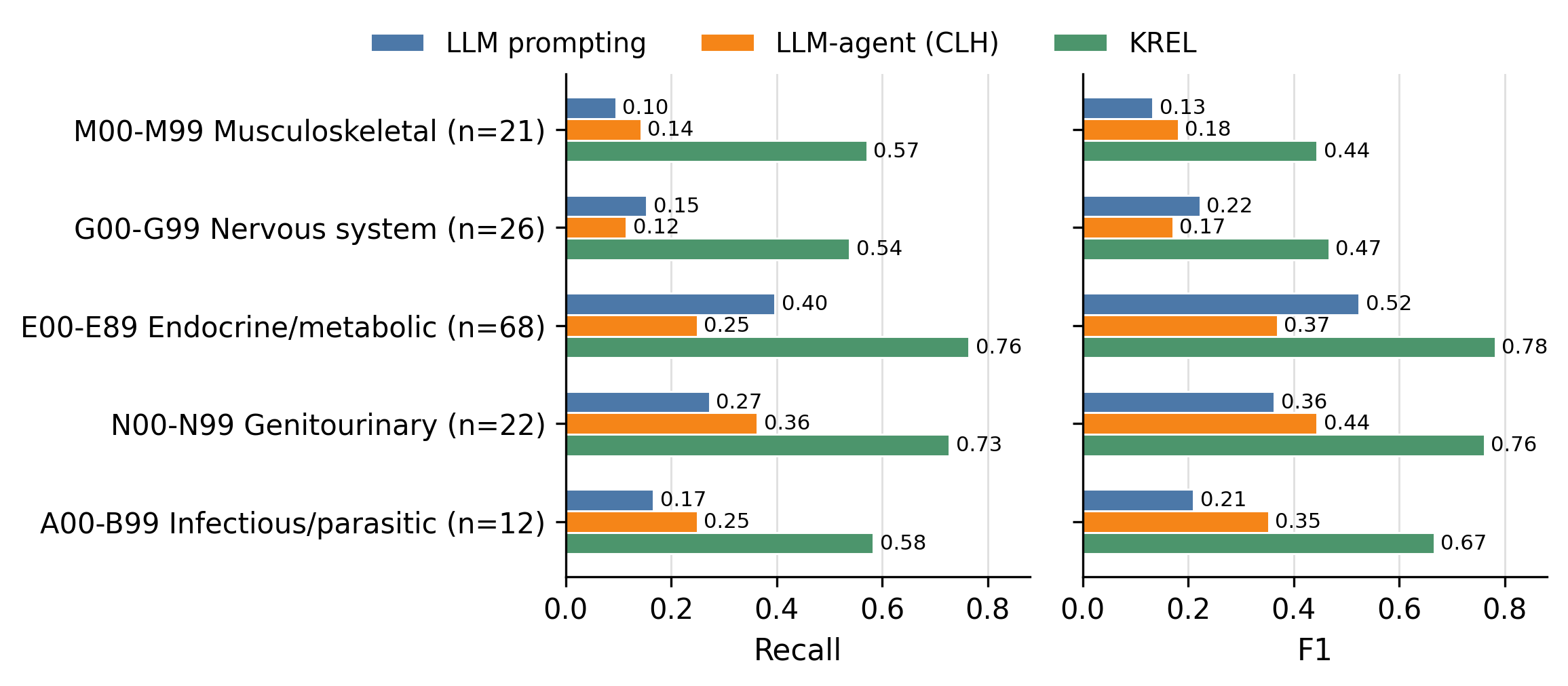}
    \caption{Performance by ICD-10-CM chapter on MDACE under the full-label-space setting. The figure compares recall and F1 for LLM prompting, the CLH LLM-agent baseline, and KREL across five ICD chapters.}
    \label{fig:chapter_method}
\end{figure*}
\subsection{Error Analysis}

Table ~\ref{tab:method_family_error_patterns_full_label} summarizes error patterns on MDACE under the full-label-space setting. KREL predicts more codes per note on average than the LLM prompting and LLM-agent baselines, with 9.67 predicted codes per note compared with 5.98 and 5.11. This leads to more false positives, but it also substantially reduces false negatives. KREL has 255 false negatives, compared with 398 for LLM prompting and 416 for the LLM-agent baseline. The LLM-agent baseline has the fewest false positives and the lowest sibling-code false-positive rate, but this pattern is partly explained by its more conservative output size. It predicts fewer codes per note and misses more gold codes.
\begin{algorithm}[!t]
\caption{Hierarchy-aware Beam Search for Candidate Recall}
\label{alg:hbs}
\KwIn{Query $q_i$, ICD hierarchy $\mathcal{G}_H =(V_C, E_H)$, scorer $r_i(c)=r_{emb}(q_i, c)$, budgets $\Theta_R=(K_c,B,M,D,K_f)$, weight $\lambda$}
\KwOut{Recalled candidate list $\mathcal{R}_i$}

Define $\mathrm{Top}_k(A,s)$ as the top-$k$ elements in $A$ ranked by score $s$\;
Define $S(x)$ as the node score\;
Initialize beam $\mathcal{B}$ with $\mathrm{Top}_{K_c}$ ICD category codes ranked by $r_i$\;
Initialize scored candidates $\mathcal{A}\leftarrow\emptyset$\;

\For{$d=1$ \KwTo $D$}{
    $\mathcal{N}\leftarrow\emptyset$\;

    \ForEach{$(p,S(p))\in\mathcal{B}$}{
        Let $\mathrm{Child}(p)$ be the child codes of $p$ in $\mathcal{G}_H$\;

        \If{$\mathrm{Child}(p)=\emptyset$}{
            Add $(p,S(p))$ to $\mathcal{A}$\;
            \textbf{continue}\;
        }

        \ForEach{$u\in \mathrm{Top}_{M}(\mathrm{Child}(p),r_i)$}{
            $S(u)\leftarrow \lambda S(p)+(1-\lambda)r_i(u)$\;
            Add $(u,S(u))$ to $\mathcal{N}$\;
            \If{$u$ is a leaf code}{
                Add $(u,S(u))$ to $\mathcal{A}$\;
            }
        }
    }

    \If{$\mathcal{N}=\emptyset$}{
        \textbf{break}\;
    }

    $\mathcal{B}\leftarrow \mathrm{Top}_{B}(\mathcal{N},S)$\;
}

$\mathcal{R}_i\leftarrow \mathrm{Top}_{K_f}(\mathcal{A},S)$\;
\Return{$\mathcal{R}_i$}\;
\end{algorithm}
KREL shows the clearest advantage in combination-code recall. It reaches 0.615, while both baselines obtain 0.077. This gap suggests that the retrieval-and-verification pipeline is more effective for codes that require information from multiple conditions or query blocks. The metric sibling-code FP rate means the number of false-positive codes that are siblings of a gold code. KREL also keeps the sibling-code false-positive rate close to the LLM-agent baseline, with 0.227 compared with 0.219, despite predicting a larger code set. This indicates that the additional predictions do not mainly come from uncontrolled sibling-code expansion.

Figure~\ref{fig:chapter_method} provides chapter-level results, indicating that the main gains are concentrated in several ICD chapters. KREL improves both recall and F1 across the five reported ICD chapters. The largest gains appear in endocrine and metabolic diseases, genitourinary diseases, nervous system diseases, musculoskeletal diseases, and infectious diseases. For example, KREL reaches 0.76 recall and 0.78 F1 in endocrine and metabolic diseases, and 0.73 recall and 0.76 F1 in genitourinary diseases. The baselines remain much lower in these chapters, especially in musculoskeletal and nervous system codes. These results indicate that KREL improves coverage in several clinically diverse chapters, rather than only increasing predictions in a single code group.

\section{Evidence Alignment Analysis}

\begin{table}[t]
\centering
\small
\begin{tabular}{lc}
\toprule
Metric & Value \\
\midrule
True-positive code pairs & 286 \\
Evidence coverage & 100\% \\
Mention-anchor coverage & 90.6\% \\
Mean semantic cosine & 0.746 \\
\bottomrule
\end{tabular}
\caption{Evidence alignment analysis on MDACE full-label-space ICD-10-CM discharge-summary evaluation. Evidence quality is evaluated only for true-positive code predictions.}
\label{tab:evidence_alignment_appendix}
\end{table}

Table~\ref{tab:evidence_alignment_appendix} evaluates whether the verifier's returned evidence aligns with human annotations on MDACE under the full-label-space setting. We compute these metrics only for true-positive code predictions, since false-positive predictions do not have corresponding gold evidence annotations. Among 286 true-positive code pairs, the verifier provides at least one evidence quote for every prediction, yielding 100\% evidence coverage. In 90.6\% of cases, at least one returned quote contains the human-annotated clinical mention for the same ICD code. The mean semantic cosine similarity between verifier-provided evidence and gold evidence is 0.746, suggesting that the returned quotes are usually close to the annotated supporting evidence, even when the exact mention is not fully matched.

Evidence coverage measures whether the verifier returns any evidence quote for a true-positive prediction. Mention-anchor coverage measures whether a returned quote contains the human-annotated clinical mention. Mean semantic cosine is computed by embedding verifier-provided evidence quotes and gold evidence sentences with \texttt{sentence-transformers/all-MiniLM-L6-v2}, and taking the maximum cosine similarity over all predicted--gold evidence pairs for each true-positive code prediction.

\section*{Generative AI Usage}

We used ChatGPT for grammar checking, language polishing, and limited coding 
support. We did not use generative AI tools to generate research ideas, experimental 
results, or references. All suggestions were reviewed and 
verified by the authors.

\section{Prompt Template}
\label{prompt_template}
\clearpage

\begin{figure*}[!htbp]
\centering
\begin{tcolorbox}[
  enhanced,
  colback=cyan!2,
  colframe=cyan!45!black,
  arc=6pt,
  boxrule=0.8pt,
  width=\linewidth,
  title=\textbf{Query Extraction Prompt},
  coltitle=black,
  colbacktitle=cyan!12,
  boxed title style={
    colframe=cyan!45!black,
    colback=cyan!12,
    arc=6pt
  },
  attach boxed title to top left={
    xshift=6pt,
    yshift*=-\dimexpr\tcboxedtitleheight/2\relax
  },
  fontupper=\small,
  top=6pt,bottom=6pt,left=8pt,right=8pt
]

You are an ICD-10-CM inpatient discharge-note diagnosis/status query extractor.
Do not output ICD codes. Output only strict JSON clinical concept objects for
candidate retrieval and later verification.

\vspace{0.5em}
\textbf{Goal:}\\
Build evidence-grounded clinical queries that preserve the documented
specificity needed for ICD-10-CM candidate recall and reranking.

\vspace{0.5em}
\textbf{Extraction policy:}
\begin{itemize}
  \item Use only provider-documented diagnoses, conditions, and clinically relevant
  statuses from the note.
  \item Do not infer diagnoses from labs, imaging, or medications alone.
  \item Include one principal reason, active conditions that affected the admission,
  and relevant chronic/history/status conditions that influenced care.
  \item Include symptoms only when no definitive diagnosis is documented for that
  problem or when the symptom is handled as a separate active problem.
  \item Preserve documented acuity, site, laterality, severity, stage, and causal
  linkage in the search query. Do not guess missing specificity.
  \item Keep status/history concepts only when explicitly documented and clinically
  relevant to the admission or discharge plan.
\end{itemize}

\vspace{0.2em}
\textbf{Section awareness:}
Scan discharge diagnoses, hospital course, assessment/plan, problem list, past
medical history, social history, and medications before finalizing the query
set. Record the section for each evidence span.

\vspace{0.2em}
\textbf{Output schema:}
\begin{lstlisting}
{
  "principal_reason": { ...one diagnosis object... },
  "active_conditions": [ ...diagnosis objects... ],
  "history_conditions": [ ...diagnosis objects... ]
}
\end{lstlisting}

\vspace{0.5em}
Output strict JSON only with no extra keys or explanatory text:

\vspace{0.25em}
\textbf{Diagnosis object schema:}
\begin{lstlisting}
{
  "base": "core clinical concept",
  "query": "search string with explicit documented modifiers/status wording",
  "modifiers": {
    "temporality": "acute"|"chronic"|"subacute"|null,
    "site": string|null,
    "linkage": [
      {"type": "with"|"without"|"due_to"|"secondary_to", "text": string}
    ]
  },
  "evidence": [
    {
      "section": "DISCHARGE_DIAGNOSES|HOSPITAL_COURSE|ASSESSMENT_PLAN|PAST_MEDICAL_HISTORY|PROBLEM_LIST|MEDICATIONS|SOCIAL_HISTORY|OTHER",
      "span": "verbatim short quote from the note"
    }
  ]
}
\end{lstlisting}

\end{tcolorbox}
\caption{Prompt for extracting and formatting queries from clinical note samples.}
\label{app:prompt:extractor}
\end{figure*}

\begin{figure*}[!htbp]
\centering
\begin{tcolorbox}[
  enhanced,
  colback=cyan!2,
  colframe=cyan!45!black,
  arc=6pt,
  boxrule=0.8pt,
  width=\linewidth,
  title=\textbf{Recall Instruction},
  coltitle=black,
  colbacktitle=cyan!12,
  boxed title style={
    colframe=cyan!45!black,
    colback=cyan!12,
    arc=6pt
  },
  attach boxed title to top left={
    xshift=6pt,
    yshift*=-\dimexpr\tcboxedtitleheight/2\relax
  },
  fontupper=\small,
  top=6pt,bottom=6pt,left=8pt,right=8pt
]

\vspace{0.2em}
\textbf{Task Instruction}\\
You are retrieving ICD-10-CM concept descriptions. 
Given a short disease/clinical concept description (often produced by an LLM), return the most semantically matching ICD-10-CM concept descriptions. Be robust to paraphrases and minor wording differences.

\end{tcolorbox}
\caption{Instruction for Qwen3-Embedding-8B}
\label{app:prompt:Qwen3}
\end{figure*}

\begin{figure*}[!htbp]
\centering
\begin{tcolorbox}[
  enhanced,
  colback=cyan!2,
  colframe=cyan!45!black,
  arc=6pt,
  boxrule=0.8pt,
  width=\linewidth,
  title=\textbf{Rerank Instruction},
  coltitle=black,
  colbacktitle=cyan!12,
  boxed title style={
    colframe=cyan!45!black,
    colback=cyan!12,
    arc=6pt
  },
  attach boxed title to top left={
    xshift=6pt,
    yshift*=-\dimexpr\tcboxedtitleheight/2\relax
  },
  fontupper=\small,
  top=6pt,bottom=6pt,left=8pt,right=8pt
]

\vspace{0.2em}
\textbf{Task Instruction}\\
You are reranking ICD-10-CM concept descriptions.
Given a clinical concept query, return the most semantically matching ICD-10-CM concepts. Prefer exact clinical meaning, preserve documented specificity, and penalize unsupported qualifiers.

\end{tcolorbox}
\caption{Instruction for Qwen3-Reranker-8B}
\label{app:prompt:Qwen3_reranker}
\end{figure*}

\begin{figure*}[!htbp]
\centering
\begin{tcolorbox}[
  enhanced,
  colback=cyan!2,
  colframe=cyan!45!black,
  arc=6pt,
  boxrule=0.8pt,
  width=\linewidth,
  title=\textbf{Rule-Aware Verification Prompt},
  coltitle=black,
  colbacktitle=cyan!12,
  boxed title style={
    colframe=cyan!45!black,
    colback=cyan!12,
    arc=6pt
  },
  attach boxed title to top left={
    xshift=6pt,
    yshift*=-\dimexpr\tcboxedtitleheight/2\relax
  },
  fontupper=\small,
  top=6pt,bottom=6pt,left=8pt,right=8pt
]

\vspace{0.2em}
\textbf{Task:}\\
Given a clinical note, evidence-grounded query blocks, and reranked ICD-10-CM
candidate descriptions, verify which candidate codes apply.

\vspace{0.5em}
\textbf{Hard constraints:}
\begin{enumerate}
  \item Do not introduce a code outside the provided global candidate set, except
   for a Stage 2 combination target explicitly listed in the note-level
   combination checks.
  \item Base decisions on the clinical note. Evidence blocks are localization hints
   and may be incomplete.
  \item Every selected code must include one or two short verbatim note quotes.
\end{enumerate}

\vspace{0.5em}
\textbf{Verdicts:}
\begin{itemize}
  \item \textbf{SUPPORTED:} explicitly documented for this admission or clearly stated in a
  clinically relevant problem/history/status context.
  \item \textbf{POSSIBLE:} clinically plausible from the note but less explicit than a
  supported diagnosis.
\end{itemize}

\vspace{0.5em}
\textbf{Rule constraint:}
Some candidate codes include normal rule hints such as "codeFirst -> X",
"useAdditionalCode -> Y", or "codeAlso -> Z". If such a rule is clinically applicable, verify the referenced rule code and include it only when supported by the note.

\vspace{0.5em}
\textbf{Stage 1: query-level candidate verification}
\begin{itemize}
  \item Verify candidates within each evidence-grounded query block.
  \item Use the clinical note, evidence spans, candidate descriptions, and normal
  coding-rule hints in the candidate block.
  \item Prefer the supported candidate with the best documented specificity.
\end{itemize}

\vspace{0.5em}
\textbf{Stage 2: across-query combination verification}
\begin{itemize}
  \item After Stage 1, review the note-level combination-rule checks.
  \item Combination codes may depend on conditions supported across different query
  blocks, so adjudicate them at the note level.
  \item Add a combination target only when it is explicitly listed in the
  combination checks and the required component code families are jointly
  supported by Stage 1 decisions or by the note.
  \item Stage 2 may change only the listed combination targets.
\end{itemize}

\vspace{0.5em}
\textbf{You are given:}\\
\textbf{The full clinical note:}\\
\texttt{\{full\_clinical note\_text\}}

\vspace{0.5em}
\textbf{Query, Evidence, Candidate code and Rule Hints for Stage 1}\\
\texttt{Query} + \texttt{Evidence} + \texttt{candidate\_code\_list\_text}+\texttt{Rule\_hints\_text}

\vspace{0.5em}
\textbf{Combination Rules for stage 2}\\
\texttt{\{combination\_rules\_text\}}

\vspace{0.5em}
\textbf{Output (STRICT JSON):} Only include SUPPORTED or POSSIBLE codes.
\begin{lstlisting}
{
  "verifications": [
    {
      "code": "I10",
      "verdict": "SUPPORTED|POSSIBLE",
      "evidence": ["verbatim quote", "verbatim quote"]
    }
  ]
}
\end{lstlisting}

\end{tcolorbox}
\caption{Prompt for rule-aware and evidence-grounded verification samples.}
\label{app:prompt:verifier}
\end{figure*}

\FloatBarrier
\clearpage

\twocolumn

\end{document}